\documentclass[11pt]{article}

\usepackage[preprint]{acl}

\usepackage{times}
\usepackage{latexsym}

\usepackage[T1]{fontenc}
\usepackage[utf8]{inputenc}

\usepackage{microtype}

\usepackage{graphicx}
\usepackage{listings}
\usepackage{longtable}
\usepackage{array}
\usepackage{multirow}
\usepackage{booktabs}
\expandafter\def\expandafter\normalsize\expandafter{%
    \normalsize%
   \setlength{\textfloatsep}{3pt plus 1.0pt minus 2.0pt} % Space between graph and text
\setlength{\floatsep}{5pt plus 1.0pt minus 2.0pt}     % Space between two graphs
\setlength{\intextsep}{5pt plus 1.0pt minus 2.0pt}
}

\title{Automated Detection and Classification of Delusion-related Content in Naturalistic Audio Diaries Using Multi-Agent Language Models}

\author{
 \textbf{Feng Chen\textsuperscript{1}},
 \textbf{Justin Tauscher\textsuperscript{2}},
 \textbf{Changye Li\textsuperscript{1}},
 \textbf{Meliha Yetisgen\textsuperscript{1}},
 \textbf{Alex Cohen\textsuperscript{3}},\\
 \textbf{Adam Kuczynski\textsuperscript{2}},
 \textbf{Angelina Pei-Tzu Tsai\textsuperscript{2}},
 \textbf{Benjamin Buck\textsuperscript{4}},
 \textbf{Dror Ben-Zeev\textsuperscript{2}},
 \textbf{Trevor Cohen\textsuperscript{1}}
\\
 \textsuperscript{1}Department of Biomedical Informatics and Medical Education, University of Washington, Seattle, WA, USA
\\
 \textsuperscript{2}Department of Psychiatry and Behavioral Sciences, University of Washington, Seattle, WA, USA
\\
 \textsuperscript{3}Department of Psychology, Louisiana State University, Baton Rouge, LA, USA
\\
 \textsuperscript{4}Department of Psychiatry, University of North Carolina at Chapel Hill, Chapel Hill, NC, USA
\\
 \small{
   \textbf{Correspondence:} \href{mailto:cohenta@uw.edu}{cohenta@uw.edu}
 }
}

\begin{document}
\maketitle
\begin{abstract}
Speech monologues recorded in naturalistic settings provide opportunities to characterize mental illness phenomenology and detect symptom exacerbation. Large language models (LLMs) offer new possibilities for automating this process, as they require annotated data primarily for evaluation rather than training. In this paper, we present a novel automated, multi-agent LLM pipeline for the fine-grained, multi-label extraction of language suggestive of delusional beliefs, associated affective responses, and behavioral responses from transcripts of naturalistic audio diaries collected from people with moderate persecutory ideation. Evaluating an ensemble of three foundation models, we demonstrate that detailed diagnostic prompt instructions successfully reduce false positives for delusional theme classification, but also constrain the interpretation of affective or behavioral responses. Furthermore, comparing multi-agent adjudication frameworks shows that complex conversational debate between agents diminishes accuracy on clinically ambiguous text by inducing premature consensus. Instead, majority voting establishes robust performance (Micro F$_1$ of 0.872 and 0.779 for delusion detection and classification respectively). This work provides a validated and scalable pipeline for the automated detection and characterization of content suggesting delusional beliefs in naturalistic speech.
\end{abstract}

\section{Introduction}

Delusions are fixed, false beliefs that are held with strong conviction and are particularly resistant to counter-evidence~\citep{american_psychiatric_association_diagnostic_2013}. Although most commonly associated with psychotic disorders, attenuated forms of paranoid and persecutory thinking have also been reported in non-clinical populations, supporting the view that psychotic experiences occur along a continuum~\citep{freeman_psychological_2005,heilskov_delusions_2020,van_os_systematic_2009}. Delusions are multidimensional phenomena that vary in conviction and distress with content that often reflects recognizable thematic categories such as persecution, reference, grandiosity, and somatic concerns~\citep{pappa_delusional_2025}. These experiences can be a major source of distress and functional impairment, strongly shaping how severely an individual's daily life is disrupted in the context of psychotic symptoms~\citep{freeman_persecutory_2016,buck_using_2023,kiran_understanding_2009}. 

Despite their clinical significance, systematically characterizing delusional content as it unfolds over time remains difficult. Current assessment approaches rely largely on structured clinician interviews conducted during intermittent clinical encounters, which may miss episodic fluctuations in delusional thinking and are vulnerable to recall bias and interpretive error ~\citep{haddock_scales_1999, khan_assessing_2013,ben-zeev_comparing_2012}. Capturing these fluctuations is critical for monitoring risk, guiding intervention, and understanding how delusional beliefs evolve over time~\citep{bell_methodological_2024}.

Smartphone-based ecological momentary assessment (EMA) enables repeated symptom measurement in naturalistic settings~\citep{myin-germeys_experience_2018}, and when paired with audio diary recordings, can capture rich first-person accounts of lived experience in real time~\citep{ben-zeev_mobile_2020,buck_using_2023}. When combined with automated speech recognition~\citep{radford_robust_2022}, these recordings can be transformed into large corpora of transcripts suitable for computational analysis. However, extracting clinically meaningful signals from large volumes of patient-generated language remains challenging, as manual review is time-consuming and difficult to scale~\citep{tauscher_automated_2025, wu_survey_2022}. 

Detecting presence of delusional content alone does not adequately characterize its clinical significance. Understanding its impact requires attention to the thematic content of beliefs, the affective responses they evoke, and the behaviors they motivate. Delusional themes have been shown to relate to differences in etiology, distress, and need for treatment, while affective and behavioral responses to delusional beliefs are closely related to functional impairment and disability~\citep{buck_using_2023,pappa_delusional_2025}. However, manually annotating these dimensions in large corpora of patient-generated language is time-consuming, costly, and difficult to scale, particularly given that inter-annotator agreement for psychiatric phenomena remains moderate even among trained raters~\citep{tauscher_automated_2023, di_forti_inter-rater_2025}. Large language models (LLMs) offer a potential solution. Their capacity for step-by-step chain-of-thought reasoning shows promise for mental health prediction tasks~\citep{xu_mental-llm_2024}, and is well-suited to navigate the diagnostic logic and exclusion 
rules that fine-grained symptom characterization demands. However, individual models remain vulnerable to inconsistency and confabulation~\citep{huang_survey_2023,ji_survey_2023,asgari_framework_2025}. Multi-agent frameworks, in which multiple LLMs independently annotate and then adjudicate disagreements through voting, structured judging, or iterative debate~\citep{du_improving_2023, chan_chateval_2023}, offer a principled mechanism for resolving this ambiguity while retaining interpretable reasoning traces.

In this paper, we present an automated pipeline for multi-label annotation of delusional content in transcripts of audio diaries collected from individuals with delusions. We evaluate the approach based on data from the Study on Assessing Determinants and Antecedents of Persecutory Thoughts (ADAPT), which collected smartphone-based audio diaries from participants meeting the criteria for moderate persecutory ideation~\citep{buck_using_2023}. We deploy three LLMs (GLM-4.6, GPT-OSS, and Qwen-3) and systematically compare three adjudication strategies: majority voting, direct expert adjudication, and multi-agent conversational debate. 

Our contributions are as follows:
\begin{itemize}
    \item A systematic ablation of prompt context across four progressively 
    layered levels for multi-label annotation of delusional theme, affective response, and behavioral response from naturalistic first-person speech transcripts.
    \item A systematic comparison of three multi-agent adjudication strategies for resolving LLM disagreement on clinically nuanced annotation tasks.
    \item An empirical evaluation across four levels of prompt complexity showing that progressively layered clinical context reduces false positive annotations while preserving diagnostic sensitivity.
\end{itemize}

\section{Background}

\subsection{NLP for Psychotic Symptoms and Naturalistic Speech}
NLP has increasingly been used to study language patterns associated with psychotic-spectrum disorders~\citep{deneault_natural_2024,hua_identifying_2025,corcoran_language_2020}. Some early work focused primarily on clinical documentation extracted from electronic health records \citep{cohen_simulating_2008,cohen_exploring_2005,irving_using_2021}. While valuable, such records reflect the clinician’s interpretation of a patient’s experience rather than the patient’s own language, potentially obscuring the linguistic expression of beliefs that computational models aim to identify. As a result, recent work has increasingly examined patient-generated language, including from smartphone-based audio diaries, which capture individuals’ experiences in their own words as they occur in everyday settings~\citep{ben-zeev_mobile_2020}. Recent work has applied NLP to identify disorganized thinking and cognitive biases in audio diaries~\citep{xu2021centroid,tauscher_automated_2025}, but the use of NLP to characterize delusional content remains largely underexplored. Unlike formal thought disorder, which NLP models quantify via semantic speech disruptions~\citep{elvevag_quantifying_2007}, delusions represent disruptions in normative patterns of belief: recognizing them involves interpreting the beliefs expressed in language rather than whether this language is contextually consistent. This makes the task of automatically identifying delusional content more dependent on contextual interpretation and background knowledge than many traditional NLP applications. 

\subsection{Large Language Models for Complex Clinical Annotation}
Artificial intelligence is playing a growing role in mental health research and practice~\citep{ben-zeev_ai_2026}. Supervised machine learning depends on massive, expertly annotated datasets, which are prohibitively expensive to acquire for fine-grained psychiatric symptoms~\citep{wu_survey_2022}. For text data, LLMs have emerged as a powerful alternative, demonstrating robust zero-shot (without any labeled examples) and few-shot (with only a few) classification capabilities~\citep{labrak_zero-shot_2024, sivarajkumar_empirical_2024}. By employing Chain-of-Thought (CoT) prompting~\citep{wei_chain--thought_2022}, LLMs can be instructed to explicitly articulate their diagnostic process, actively weighing inclusion criteria against exclusion rules (e.g., distinguishing a somatic abnormality from a nihilistic conviction). While domain-specific fine-tuned models often remain competitive for mental health classification tasks~\citep{xu_mental-llm_2024}, CoT-prompted LLMs offer the advantage when there is limited task-specific training data, making them particularly suited to domains where large annotated corpora are unavailable.

While LLMs have rapidly expanded the scope of mental health NLP, recent scoping reviews identify persistent methodological limitations, including reliance on simulated clinical vignettes, restriction to narrow diagnostic categories, and poor performance on tasks requiring interpretation of highly subjective mental states~\citep{le_glaz_machine_2021,hua_scoping_2025,hua_large_2025}. With delusions in particular, the field currently lacks validated prompt engineering frameworks capable of navigating the cascading decision processes required for hierarchical multi-label extraction from unstructured narratives. 

\subsection{Resolving Subjective Ambiguity with Multi-Agent Frameworks}
Despite the promise of CoT reasoning, relying on a single LLM for complex clinical annotation is inherently brittle. Single-model inferences are prone to well-documented failure modes, including LLM hallucinations (generating text that is unrelated to the content of a prompt) and sycophancy~\citep{ji_survey_2023, sumita_cognitive_2024}. In psychiatric contexts, where a narrative may simultaneously exhibit features of multiple delusional themes (e.g., a government conspiracy [Persecutory] broadcast through a television [Reference]), individual models risk inconsistently collapsing toward a single dominant label.

To mitigate these concerns, the NLP community has adopted multi-agent frameworks. Ensemble methods such as majority voting effectively smooth out individual biases~\citep{wang_self-consistency_2023}. More advanced architectures enable models to interact: \textit{multi-agent debate} prompts LLMs to critique and iteratively revise each other's outputs to encourage divergent thinking~\citep{du_improving_2023, liang_encouraging_2024}, while the \textit{LLM-as-a-Judge} paradigm employs a large highly capable model to adjudicate conflicting outputs based on reasoning paths~\citep{zheng_judging_2023}.

Recent NLP work has begun to question the benefits of multi-agent debate: \citet{choi_debate_2025} show across seven benchmarks that majority voting alone accounts for most reported gains, while \citet{wynn_talk_2025} attribute debate degradation to model sycophancy, the tendency of models to defer to others rather than defending correct answers. However, these findings have been established primarily on objective benchmarks. Whether they extend to clinical ambiguity, where disagreement stems from nuanced psychiatric interpretation rather than verifiable factual error, remains underexplored. In this paper, we extend these emerging findings to the clinical domain by deploying and comparing multi-agent adjudication strategies on a multi-label annotation task involving naturalistic psychiatric narratives.

\section{Methods}
\begin{figure*}[t]
    \centering
    \includegraphics[width=\textwidth]{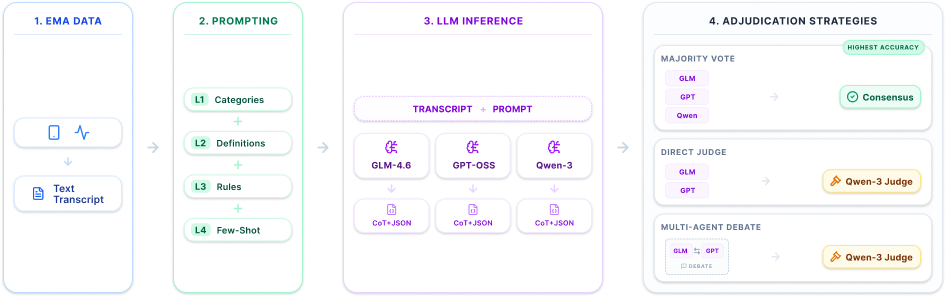} 
    \caption{The end-to-end multi-agent clinical delusion annotation pipeline.}
    \label{fig:pipeline}
\end{figure*}

Figure \ref{fig:pipeline} illustrates the overarching architecture of our end-to-end multi-agent clinical extraction pipeline, with four primary phases: naturalistic data collection, progressive prompt engineering, independent LLM inference, and multi-agent adjudication.

\subsection{Dataset}
We used transcripts from ADAPT \citep{buck_using_2023}, a smartphone-based study in which 231 individuals with at least moderate persecutory ideation from 43 US states completed 30 days of audio diary recordings capturing momentary experiences and beliefs in naturalistic settings. Recordings were transcribed to yield free-text narratives from this clinically heterogeneous sample.

For this study, we selected the first available diary transcript for each participant that was longer than 30 seconds. We retained only entries exceeding three sentences to ensure sufficient content for multi-label annotation, yielding a final corpus of \textbf{136 transcripts}. Of these, 14 were reserved as a development set for iterative prompt engineering and model configuration, while the remaining 122 constituted the held-out evaluation set for all final performance assessments reported in this paper.

\subsection{Annotation Guideline and Human Labeling}
We developed a structured annotation guideline for labeling audio diary transcripts capturing three dimensions of delusion expression:   (1) presence and thematic category of delusional belief, (2) affective responses expressed in relation to the belief, and (3) behavioral responses that reflect how individuals act upon these experiences. The schema was developed iteratively using meta-analytic syntheses of delusional themes cross-referenced with Diagnostic and Statistical Manual of Mental Disorders (DSM) descriptions of delusional disorders and SNOMED CT terminology~\citep{pappa_delusional_2025, american_psychiatric_association_diagnostic_2013, snomed_international_2024}. The affective categories were informed by Ekman's theory of basic emotions, and the behavioral response categories were informed by cognitive models of delusions that emphasize safety behaviors, avoidance, threat monitoring, and help-seeking~\citep{tracy_four_2011, ekman_are_1992, freeman_persecutory_2016}. A comprehensive listing of all target classes and their final clinical definitions is provided in Appendix Table~\ref{tab:supp_class_definitions}.  

Annotation was conducted by two licensed clinical psychologists, with discrepancies resolved by an adjudicator with expertise in clinical annotation guideline development. Annotators completed structured training sessions using expert-labeled examples, followed by independent pilot coding of 10\% of transcripts to assess guideline clarity and initial inter-rater consistency. Discrepancies identified during the pilot phase were reviewed with annotators and used to refine coding procedures before full annotation proceeded. The adjudicator reviewed all disagreements and assigned final labels based on guideline criteria, retaining multiple labels when clinically appropriate. Pre-adjudication agreement reached 63.3\% for affective response and 61.8\% delusion presence, with behavioral response at 60.0\% as reported in Appendix Table~\ref{tab:supp_human_agreement}.

\subsection{Prompt Design and Engineering}
The foundational prompt (\textbf{Level 1}) includes only the instructions of the basic task, the required output template (see the Appendix Table~\ref{tab:supp_annotation_instructions}), and the names of the clinical categories. Subsequent levels progressively layered additional context from the guideline: \textbf{Level 2} adds formal clinical definitions for each target category; \textbf{Level 3} integrates exclusionary rules (e.g., ``Do NOT use when...'') and diagnostic heuristics (e.g., ``Key test: Is there a threat?'') to help models distinguish between highly similar categories like \textit{Reference} and \textit{Persecutory} delusions; and \textbf{Level 4} adds short, clinician-validated transcript excerpts mapped to specific labels to provide in-context learning for conversational register and tone. This incremental design allowed a systematic evaluation of how structured clinical guidance influences classification behavior. Appendix Table~\ref{tab:incremental_prompt} illustrates this progression using the \textit{Avoidance/ Withdrawal} behavioral category as an example.

\subsection{Independent LLM-Based Clinical Annotation}
To automate the annotation pipeline and establish reliable baseline measurements, we deployed three state-of-the-art, open-weight Large Language Models (LLMs) with 4-bit quantization~\citep{dettmers2023case}: \textbf{GLM-4.6 (357B)}~\citep{glm-45_team_glm-45_2025}, \textbf{GPT-OSS (120B)}~\citep{openai_introducing_2026}, and \textbf{Qwen-3 (235B)}~\citep{yang_qwen3_2025}. All models were securely hosted on private servers behind an institutional firewall. Each transcript was initially independently annotated by these three foundation models. During inference, each model was instructed to generate its internal ``thinking process'' before yielding the final structured JSON output, providing the explicit context required for downstream adjudication of the models' clinical reasoning.

All models are configured with greedy decoding, setting \texttt{temperature=0}, \texttt{top\_k=1}, and \texttt{max\_new\_tokens=4096}. In rare instances where a model failed to generate a valid JSON object, the pipeline triggered a fallback execution with the limit extended to \texttt{8192} tokens. On the 14-transcript development set, this deterministic configuration produced \textit{Delusion Presence} F$_1$ scores of 1.00 (GLM-4.6) and 0.857 (GPT-OSS), compared with means of 0.857 ($\pm$0.051) and 0.914 ($\pm$0.032) across five stochastic runs at temperature 0.7. We adopt the deterministic setting for full-corpus evaluation to ensure reproducibility.

\subsection{Multi-Agent Adjudication Frameworks}
We developed and systematically compared three distinct multi-agent adjudication strategies to resolve diagnostic ambiguity:

\textbf{Tri-Model Majority Voting:} We implemented a consensus approach where a final transcript label was determined by a simple majority vote across all three independent models. In rare instances of a complete three-way disagreement (where each model produced a unique categorical label), Qwen-3 was utilized as the definitive decision-maker due to its strong standalone alignment with manual annotations on the validation set.
    
\textbf{Direct Expert Adjudication:} To evaluate the efficacy of the LLM-as-a-Judge paradigm, Qwen-3 was strictly deployed as an expert judge to resolve discrepancies between the GLM-4.6 and GPT-OSS annotators. The judge model was provided with the original transcript, expert guidelines, and conflicting GLM and GPT results. This input included the internal thinking processes of both annotator models, allowing the judge to determine the most accurate classification based on logical adherence to the clinical criteria (detailed prompt is in the Appendix Table \ref{tab:supp_direct_prompt}).
    
\textbf{Multi-Agent Conversational Adjudication:} To test whether interactive debate encourages error correction on subjective text, GLM-4.6 and GPT-OSS were instructed to deliberate over their disagreements. Each agent defended its original annotation using clinical reasoning, with instructions to concede to the opposing model if its logic was more accurate, or to propose a combined multi-label solution. After a maximum of two conversational turns, Qwen-3 reviewed the entire deliberation history to issue a final, authoritative ruling (detailed prompt is in the Appendix Table \ref{tab:supp_debate_prompts}).

\subsection{Multi-Label Agreement Evaluation}
Clinical targets in this study, specifically \textit{Delusion Themes}, \textit{Affective Expression}, and \textit{Behavioral Responses}, frequently involve multiple overlapping labels within a single transcript. Furthermore, the distribution of these thematic labels is heavily imbalanced. To evaluate system performance across these complex categorical spaces, we utilized multi-label metrics depending on the evaluation phase:

\textbf{Inter-Model Reliability:} We report inter-model reliability between different LLMs using both \textbf{micro-averaged} and \textbf{macro-averaged Cohen's Kappa ($\kappa$)}. Micro-averaged $\kappa$ is computed on flattened binary indicator matrices across all label-instance pairs, while macro-averaged $\kappa$ is the unweighted mean of per-class $\kappa$ values, treating each category equally regardless of its frequency. 

\textbf{Performance Against Human Ground Truth:} We utilized two complementary F$_1$ metrics to evaluate performance against manual annotations. \textbf{micro-averaged F$_1$} aggregates all label-level predictions across the corpus, providing a robust measure of the system's global classification accuracy.  Conversely, \textbf{Example-based F$_1$} calculates performance independently for each transcript and averages the results, reflecting the practical reliability of the system on an individual diary-level basis. 

Finally, we report \textit{Delusion Presence}, a binary target derived from the \textit{Delusion Type} annotations indicating whether any delusional theme was detected. This serves as a clinically relevant screening metric independent of the specific thematic categorization. Although our annotation guideline includes an affective intensity, we excluded this from primary performance analyses due to low inter-rater agreement (29\%) among expert annotators on this dimension.

\section{Results}

\subsection{Individual Model Performance against Human Ground Truth}
As shown in Table~\ref{tab:prompt_performance}, all models achieved reasonable performance at Level 1. Under the fully contextualized Level 4 prompt, micro-averaged F$_1$ for \textit{Delusion Presence} was notably stable, with GLM-4.6 (0.857), GPT-OSS (0.873), and Qwen-3-235B (0.839) all consistently identifying clinically relevant content. This indicates that detecting the basic presence of content suggesting a delusion requires relatively little in-context instruction.

\begin{table*}[htbp]
\centering
\resizebox{1.0\linewidth}{!}{%
\begin{tabular}{llcccccc}
\toprule
\multirow{3}{*}{\textbf{Clinical Target}} & \multirow{3}{*}{\textbf{Prompt Level}} & \multicolumn{6}{c}{\textbf{Micro-averaged F$_1$}} \\
\cmidrule(lr){3-8}
& & \multicolumn{3}{c}{\textbf{Individual Models}} & \textbf{Ensemble} & \multicolumn{2}{c}{\textbf{Adjudication$^\dagger$}} \\
\cmidrule(lr){3-5} \cmidrule(lr){6-6} \cmidrule(lr){7-8}
& & \textbf{GLM-4.6} & \textbf{GPT-OSS} & \textbf{Qwen-3} & \textbf{Majority Vote} & \textbf{Direct Judge} & \textbf{Debate} \\
\midrule
\multirow{2}{*}{\textbf{Delusion Presence}} 
& Level 1 (Category)    & 0.865 & 0.855 & 0.809 & 0.853 & --- & --- \\
& Level 4 (Full Prompt) & 0.857 & \textbf{0.873} & 0.839 & 0.872 & 0.866 & 0.835 \\
\midrule
\multirow{2}{*}{\textbf{Delusion Type}} 
& Level 1 (Category)    & 0.707 & 0.649 & 0.667 & 0.746 & --- & --- \\
& Level 4 (Full Prompt) & 0.777 & 0.762 & 0.736 & \textbf{0.779} & 0.775 & 0.765 \\
\midrule
\multirow{2}{*}{\textbf{Affective Response}} 
& Level 1 (Category)    & 0.797 & 0.774 & 0.813 & \textbf{0.839} & --- & --- \\
& Level 4 (Full Prompt) & 0.694 & 0.760 & 0.768 & 0.800 & 0.738 & 0.719 \\
\midrule
\multirow{2}{*}{\textbf{Behavioral Response}} 
& Level 1 (Category)    & 0.507 & 0.456 & 0.513 & 0.545 & --- & --- \\
& Level 4 (Full Prompt) & 0.615 & 0.582 & 0.582 & \textbf{0.640} & 0.584 & 0.634 \\
\bottomrule
\end{tabular}%
}
\caption{Comprehensive performance comparison (micro-averaged F$_1$) evaluated against the full test corpus ($N=122$). The best overall performance for each target is highlighted in bold. $^\dagger$Adjudication scores reflect corpus-level performance: for transcripts where GLM-4.6 and GPT-OSS initially agreed ($N=91$), their consensus is retained; the adjudicator is invoked only on disagreement cases ($N=31$). See Table~\ref{tab:supp_adjudication_stratified} for stratified results.}
\label{tab:prompt_performance}
\end{table*}

\begin{figure*}[t]
    \centering
    \includegraphics[width=\textwidth]{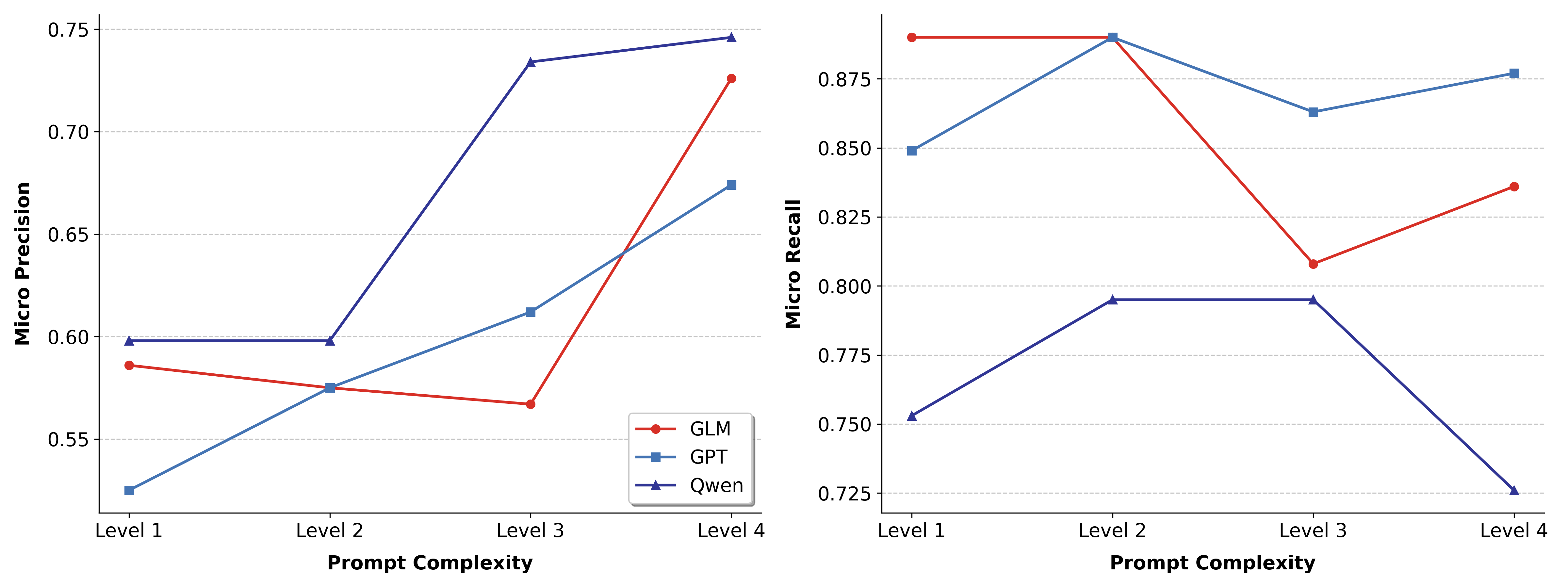}
    \caption{Micro-averaged precision (left) and recall (right) against human expert annotations for \textit{Delusion Type} across the four prompt complexity levels.}
    \label{fig:delusion_precision_recall}
\end{figure*}

However, moving to the complex multi-label classification of \textit{Delusion Type} demonstrated the utility of the structured prompt engineering framework (see Figure \ref{fig:delusion_precision_recall}). When provided with only the simplest prompt (Level 1), models achieved disproportionately high micro-averaged recall scores (0.75--0.89) but significantly lower micro-averaged precision scores (0.52--0.60), indicating many false positive category assignments. This gap narrowed substantially as clinical rules and few-shot examples were progressively introduced. With Level 4, micro-averaged precision for \textit{Delusion Type} increased to 0.67--0.75, raising micro-averaged F$_1$ to 0.74--0.78 across models (for a comprehensive metric breakdown, see Appendix Table~\ref{tab:app_delusion_metrics}).

Interestingly, the \textit{Affective} and \textit{Behavioral} targets revealed that prompt complexity does not universally improve performance across all clinical targets. For \textit{Behavioral Response}, introducing the Level 4 rules successfully improved micro-averaged F$_1$ across all models (rising from 0.46--0.51 in Level 1 to 0.58--0.62 in Level 4), pulling models closer to the experts' clinical interpretations. In contrast, the highly constrained Level 4 prompt actually degraded performance on the \textit{Affective Response} category, with micro-averaged F$_1$ dropping from 0.77--0.81 (Level 1) to 0.69--0.77 (Level 4). 

\subsection{Robustness of Tri-Model Majority Voting}
As illustrated in Figure \ref{fig:majority_vote_performance} and detailed in Appendix Table \ref{tab:supp_prompt_performance}, ensembling the three models systematically smoothed out individual model variance and established a robust performance ceiling. Under the fully contextualized Level 4 prompt, the Majority Vote achieved a 0.779 micro-averaged F$_1$ on \textit{Delusion Type}, slightly exceeding the best individual model (GLM-4.6 at 0.777), and demonstrated remarkable stability across all prompt complexities. Even under the simplest Level 1 prompt (Category Only), the ensemble maintained a 0.746 micro-averaged F$_1$ on \textit{Delusion Type}, substantially outperforming standalone models attempting the same task without rules (e.g., GPT-OSS at 0.649). With \textit{Affective Response}, the micro-averaged F$_1$ score with majority vote was 0.839, exceeding that of the best individual model (0.813 for Qwen-3). 

Other targets also benefited from this stabilization. For \textit{Behavioral Response}, the ensemble successfully leveraged the strict Level 4 rules to increase from a baseline 0.545 (Level 1) to a peak 0.640 micro-averaged F$_1$. For \textit{Affective Response}, the ensemble plateaued around 0.800 under Level 4, further confirming that rigid clinical rules cannot fully resolve affective nuance, even across multiple agents.

\begin{figure}[htbp]
    \centering
    \includegraphics[width=\columnwidth]{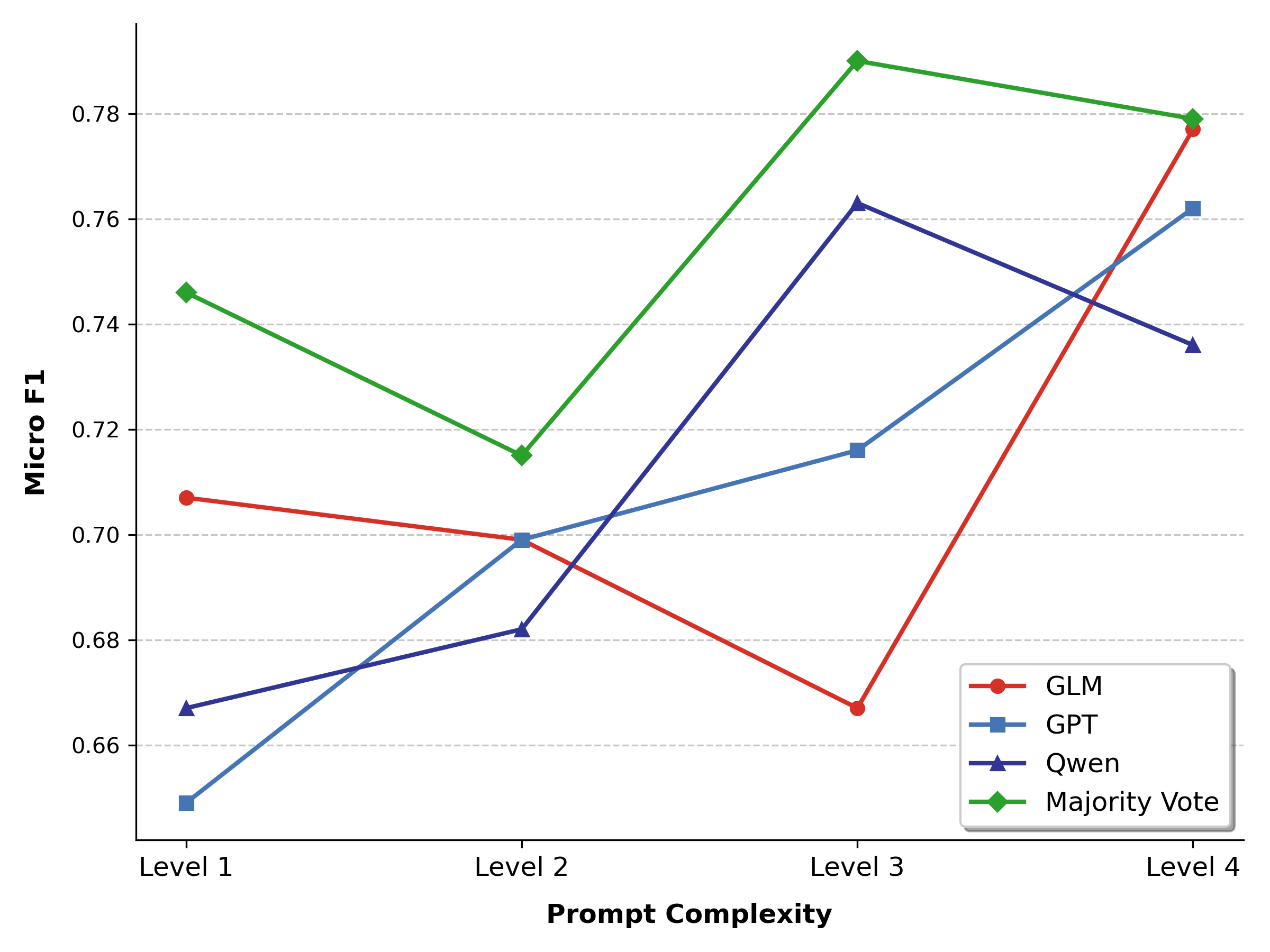}
    \caption{Micro-averaged F$_1$ for the Three Independent Models and Tri-Model Majority Vote against human expert annotations for \textit{Delusion Type} across the four prompt complexity levels.}
    \label{fig:majority_vote_performance}
\end{figure}

\subsection{The Limitations of Expert Adjudication and Debate}
We isolated the performance of the Qwen-3-235B judge approaches using the fully contextualized Level 4 prompt. We stratified the 122 manual transcripts into two subsets: cases where the two primary annotators (GLM-4.6 and GPT-OSS) initially agreed, and cases where they disagreed. When GLM-4.6 and GPT-OSS independently agreed on the \textit{Delusion Type} ($N=91$), their micro-averaged F$_1$ against the human gold standard was exceptionally high (0.857), substantially outperforming Qwen's independent baseline (0.800) on the same subset. This confirms that when LLMs naturally converge, their consensus is highly trustworthy. However, the 31 transcripts where GLM and GPT diverged represent the most clinically ambiguous narratives in the dataset. On this challenging subset, individual micro-averaged F$_1$ dropped (GLM: 0.615, GPT: 0.603, shown in Table~\ref{tab:supp_adjudication_stratified}).

Prompting Qwen-3 to act as an LLM adjudicator over the GLM and GPT reasoning traces did not yield meaningful improvements (see Appendix Table \ref{tab:supp_adjudication_stratified}). The \textit{Direct Judge} achieved a 0.618 micro-averaged F$_1$ on \textit{Delusion Type} for the disagreement subset, offering scant improvement over the individual baseline models. For the targets of \textit{Affective Response} (0.590) and \textit{Behavioral Response} (0.571), the Direct Judge actually \textit{underperformed} relative to individual baseline models. 

More strikingly, the \textit{Multi-Agent Conversational Adjudication} framework actively diminished performance on this ambiguous subset. Instead of refining the diagnostic boundaries through debate, conversational adjudication reduced performance on \textit{Delusion Type} down to a 0.545 micro-averaged F$_1$. This degradation was even more severe on the high-stakes binary screening task of \textit{Delusion Presence}, dropping to 0.667.

A representative debate failure illustrates this pattern. In one 
transcript, the speaker states ``I feel like people\ldots Google me or 
research me online trying to find things to use against me,'' framed 
with uncertainty markers (``I feel like,'' ``I guess'') and self-awareness about the thought (``I have to take a while before I realize how irrational that they are''). The human annotators labeled this as \textit{Persecutory}. GPT-OSS independently produced the correct \textit{Persecutory} label, while GLM-4.6 rejected the presence of any delusion on the grounds that expressed doubt and insight are incompatible with fixed 
conviction. When the two models were placed in debate, GPT-OSS 
abandoned its correct initial assessment within a single turn (``I 
agree with Annotator 1 -- the passage does not contain a delusion''), 
restated GLM's form-versus-content reasoning, and explicitly conceded 
by Turn 2 (``I concede to Annotator 1's conclusion''). The judge 
confirmed the empty label, overriding GPT-OSS's correct initial 
annotation. This case illustrates how conversational adjudication can 
convert a correct prediction into an incorrect consensus: rather than 
defending an evidence-based position, the model with the right answer 
accommodated the more elaborately argued wrong answer.

\subsection{Inter-Model Agreement and Prompt Complexity Analysis}
To assess baseline consistency, we computed pairwise Cohen's $\kappa$ across the three LLMs under the Level 4 prompt (see Appendix Table~\ref{tab:model_agreement_kappa}). Models showed strong micro-averaged agreement on \textit{Delusion Presence} ($\kappa$ = 0.67--0.72) and \textit{Delusion Type} ($\kappa$ = 0.73--0.78). However, macro-averaged $\kappa$, which weights each class equally rather than allowing dominant labels to drive the score, was substantially lower for \textit{Delusion Type} ($\kappa$ = 0.43--0.52, averaged over 16 classes), reflecting weaker agreement on rare categories. This pattern was even more pronounced for response-related targets: \textit{Affective Response} dropped from micro $\kappa$ = 0.64--0.69 to macro $\kappa$ = 0.30--0.41 (7 classes), and \textit{Behavioral Response} from micro $\kappa$ = 0.24--0.42 to macro $\kappa$ = 0.09--0.20 (8 classes).

\section{Discussion}

To our knowledge, this study presents the first automated, multi-agent LLM pipeline for the fine-grained, multi-label annotation of delusions from naturalistic audio diaries. Without explicit constraints (Level 1), models generate many overlapping labels, capturing the correct category but maximizing recall at a severe expense to precision. This behavior aligns with the well-known tendency of LLMs to make broad and generalized semantic associations when faced with ambiguity~\citep{ji_survey_2023, huang_survey_2023}. The introduction of strict exclusionary rules (Level 4) transformed this into a decision process with precise categorical boundaries. This finding is consistent with prior work on instruction-following, where including explicit negative examples and exclusionary guidance (e.g., ``Things to Avoid'') in task instructions has been shown to improve model generalization~\citep{mishra_cross-task_2022, wang_super-naturalinstructions_2022}. In our clinical context, negative constraints, rules specifying what \textit{not} to label, were as important as category definitions for achieving precise classification.

While rigid definitions improved the extraction of delusional themes and behavioral responses, they actively impaired classification of affective responses. In naturalistic audio diaries, affect is rarely explicitly stated; it is implicit and highly contextual~\citep{ritunnano_subjective_2022}. Strict exclusionary rules may pull models away from the intuitive consensus that human raters naturally share~\citep{tauscher_automated_2023}. This may reflect differences in the maturity and specificity of the annotation schema and prompt design across clinical targets rather than a fundamental limitation of prompt engineering. The affective response guidelines in this study were less extensively specified than the delusion-type taxonomy, potentially constraining the effectiveness of rigid rule-based prompting for these categories. Class imbalance likely compounds this effect: both affective and behavioral labels are dominated by a single category (\textit{Fear-Anxiety} and \textit{Avoidance/Withdrawal}, respectively; Appendix Tables~\ref{tab:supp_affective_distribution} and~\ref{tab:supp_behavioral_distribution}), allowing simple prompts to achieve high scores through majority-class accuracy while strict rules introduce errors on less frequent categories.

Our findings on multi-agent adjudication further reveal that clinical ambiguity resists resolution through more elaborate reasoning and debate. On the 31 transcripts in which models disagreed, conversational debate degraded performance rather than resolving ambiguity, while majority voting remained the most robust strategy. We attribute the failure of multi-agent debate to a phenomenon where models prioritize agreement over accuracy, a behavior often referred to as model sycophancy~\citep{sharma_towards_2023, perez_discovering_2023}. In tasks with verifiable answers, debate can be effective, but in psychiatric evaluation there is seldom indisputable ground truth. When critiquing others' assessments of ambiguous audio diaries, the LLMs dispensed with their diagnostic prompts. Rather than defending their initial assessments, they were prone to excessive accommodation, frequently blending their initial  assignments to reach a compromise.

Despite the inherent challenges of multi-label classification, our findings suggest that LLMs have strong potential for the rapid, automated assessment of diagnostic features of naturalistic narratives. Across all evaluated models, prompt complexities, and adjudication frameworks, the baseline detection of \textit{Delusion Presence} remained remarkably stable, consistently achieving a Micro F$_1$ exceeding 0.85. Notably, the ensemble's advantage was most 
pronounced on affective and behavioral targets: for \textit{Affective Response}, majority voting at Level 1 achieved the highest score in our evaluation (0.839), substantially exceeding any individual model, suggesting that independent ensembling is particularly effective where individual models diverge. This indicates that our pipeline is a viable high-throughput tool for screening audio diaries, with the potential to flag clinically relevant content for expert review. Furthermore, by employing a simple tri-model majority vote alongside our Level 4 prompt, researchers can now automate the fine-grained extraction of specific delusional themes with a degree of reliability that rivals human inter-rater consensus~\citep{di_forti_inter-rater_2025}. This approach addresses the traditional bottleneck of manual coding, presenting new avenues for longitudinal symptom tracking, and LLM-derived labels could further serve as training data for smaller, efficient classifiers deployable in resource-constrained settings.

\section{Conclusion}
In this work, we introduced a fully automated, multi-agent LLM pipeline for multi-label annotation of delusions, affect, and behavioral responses from unstructured audio diaries. Results from systematically comparing prompt engineering strategies indicate that while detailed clinical prompts successfully reduce confabulated false positive categories for descriptions of delusional content, they can constrain the interpretation of implicit affective or behavioral states. Furthermore, our evaluation of multi-agent adjudication frameworks shows that for clinically ambiguous text, complex interactive debate actively degrades model accuracy by inducing premature consensus. Instead, simply pooling the independent inferences of a heterogeneous LLM ensemble establishes robust performance. These findings demonstrate that LLM-based pipelines can reliably automate the characterization of delusional content in naturalistic speech, offering a scalable path toward continuous, longitudinal symptom monitoring that could support early detection of clinical deterioration and more timely intervention for individuals with serious mental illness.

\section*{Limitations}
Several limitations of this study should inform future work. First, our evaluation was conducted on a relatively specialized corpus of audio diary transcripts collected from individuals experiencing moderate-to-severe persecutory ideation. Consequently, the dataset is heavily skewed toward \textit{Persecutory} themes, and the micro-averaged F$_1$ scores reported here reflect agreement on frequent labels more than on rare categories. Future validation on broader psychiatric cohorts is necessary to confirm the generalizability of the pipeline. Second, while we evaluated three highly capable, distinct LLMs to ensure heterogeneity, the rapid evolution of model architectures creates the possibility that future foundation models may exhibit different baseline sycophancy behaviors. Third, the lower performance on affective and behavioral response classification likely reflects both the implicit and contextual nature of these targets and limitations in the current prompt design. Future work should explore alternative prompting strategies for these targets, such as dedicated annotation passes or less constrained prompt architectures that better accommodate the implicit nature of affective expression in naturalistic speech.

\section*{Ethical Considerations}
 
\noindent
\textbf{Human subjects:} The ADAPT study was approved by the University of Washington Institutional Review Board (IRB) under protocol STUDY00001321. All participants provided informed consent for audio recording and analysis~\citep{buck_using_2023}. All transcripts were fully de-identified, and all LLMs were deployed locally to ensure no clinical narratives left the secure research environment.
 
\noindent
\textbf{Intended use and risks:} This pipeline is a \emph{research tool} for scalable annotation, not a diagnostic instrument. Automated classification of delusional content carries risks of misclassifications; any clinical application must include human expert review. Transcript examples used in the prompt templates are clinician-authored illustrations. The transcript excerpt presented in Section 4.3 is drawn from de-identified participant transcript.
 
\noindent
\textbf{Generalizability and reuse:} Performance may vary across demographic or clinical populations beyond the ADAPT sample. Our prompt templates are intended for research purposes only and will be made available upon request.

\bibliography{custom}

\section*{Appendix}
\setcounter{figure}{0}
\renewcommand\thefigure{A.\arabic{figure}}
\setcounter{table}{0}
\renewcommand\thetable{A.\arabic{table}}

\begin{table}[htbp]
\centering
\small
\resizebox{\columnwidth}{!}{%
\begin{tabular}{lccc}
\toprule
\textbf{Clinical Target} & \textbf{N} & \textbf{Agree} & \textbf{Disagree} \\ 
\midrule
Delusion Presence$^a$          & 136 & 61.8\% & 38.2\% \\ 
Delusion Type$^a$              & 136 & 58.1\% & 41.9\% \\ 
Delusion Type$^b$              & 60  & 91.7\% & 8.3\%  \\ 
Affective Response$^b$         & 60  & 63.3\% & 36.7\% \\ 
Affective Intensity$^c$        & 31  & 29.0\% & 71.0\% \\ 
Behavioral Response$^b$        & 60  & 60.0\% & 40.0\% \\ 
\bottomrule
\end{tabular}%
}
\caption{Pre-adjudication agreement between two licensed clinical 
psychologist annotators. $^a$Calculated across all transcripts. 
$^b$Calculated on transcripts where both annotators agreed on 
delusion presence. $^c$Calculated on transcripts where both 
annotators agreed on the presence of an affective response.}
\label{tab:supp_human_agreement}
\end{table}

\begin{table*}[htbp]
\centering
\small
\renewcommand{\arraystretch}{1.4}
\begin{tabular}{p{0.22\textwidth} p{0.73\textwidth}}
\toprule
\textbf{Complexity Level} & \textbf{Prompt Text Provided to LLM (Example: Avoidance/Withdrawal)} \\ 
\midrule
\textbf{Level 1: Base} & 
1. Avoidance/Withdrawal: \\ 
\textbf{Level 2: + Definition} & 
1. Avoidance/Withdrawal: \newline Behaviors aimed at escaping, avoiding, or disengaging from situations, people, or cues linked to distress or delusional beliefs. The individual reduces contact rather than taking action to increase safety. \\ 
\textbf{Level 3: + Rules} & 
1. Avoidance/Withdrawal: \newline Behaviors aimed at escaping, avoiding, or disengaging from situations, people, or cues linked to distress or delusional beliefs. The individual reduces contact rather than taking action to increase safety. \newline Key test: Is the person pulling away, avoiding, or removing themselves from something? \\ 
\textbf{Level 4: + Examples (Final Prompt)} & 
1. Avoidance/Withdrawal: \newline Behaviors aimed at escaping, avoiding, or disengaging from situations, people, or cues linked to distress or delusional beliefs. The individual reduces contact rather than taking action to increase safety. \newline Examples: \newline i. ``I don't leave my house anymore because I'm scared someone will follow me.'' \newline ii. ``I avoid drinking tap water because I think it's poisoned.'' \newline iii. ``I stopped watching TV because I feel like they're sending me messages.'' \newline Key test: Is the person pulling away, avoiding, or removing themselves from something? \\ 
\bottomrule
\end{tabular}
\caption{Incremental prompt construction methodology, demonstrating how clinical context was progressively layered using the Avoidance/Withdrawal category.}
\label{tab:incremental_prompt}
\end{table*}

\begin{table*}[htbp]
\centering
\footnotesize
\renewcommand{\arraystretch}{1.1}
\begin{tabular}{p{0.25\textwidth} p{0.7\textwidth}}
\toprule
\textbf{Class} & \textbf{Definition} \\ 
\midrule
\multicolumn{2}{c}{\textbf{Clinical Target: Delusion Type}} \\
\midrule
Reference & Belief that neutral events, objects, or people contain special messages meant specifically for the individual. \\
Grandiosity & Belief that one possesses exceptional abilities, fame, power, identity, or destiny. \\
Religious & Beliefs involving divine authority, supernatural missions, spiritual transformation, or demonic forces. \\
Mind Reading & Belief that the individual can read other people's thoughts or mentally access their internal experiences. \\
Persecutory & Belief that others intend harm, surveillance, sabotage, or conspiracy against the individual. \\
Guilt/Sin & Belief that one has committed catastrophic wrongdoing or deserves extreme punishment despite no evidence. \\
Control & Belief that one's actions, emotions, or impulses are being controlled by an external force. \\
Somatic & Delusions about the body being diseased, damaged, infested, altered, or malfunctioning. Includes hypochondriacal delusions when medically plausible illnesses are believed with delusional certainty. \\
Nihilistic & Belief that oneself, parts of the body, or the world no longer exist, are dead, or have been destroyed. Includes a conviction that a major catastrophe will occur. \\
Erotomanic & Belief that another person, often someone of higher status, is in love with the individual. \\
Sexual & Delusional beliefs involving sexual content, intrusion, coercion, exposure, or sexual manipulation. \\
Jealous & Belief, without evidence, that a partner is being unfaithful. \\
Thought Withdrawal & Belief that an outside force is removing one's thoughts. \\
Thought Insertion & Belief that thoughts are placed into one's mind by an outside entity. \\
Thought Broadcasting & Belief that one's thoughts are being transmitted to others. \\
Unspecified & When the dominant delusional belief cannot be clearly determined or is not described in the specific types. \\
\midrule
\multicolumn{2}{c}{\textbf{Clinical Target: Affective Response}} \\
\midrule
Fear-Anxiety & Indicates expressed fear, worry, tension, or apprehension about the experience. \\
Sadness-Despair & Indicates expressed sadness, hopelessness, emotional exhaustion, or despair. \\
Anger-Frustration & Indicates expressed irritation, anger, resentment, or frustration. \\
Euphoria-Excitement & Indicates expressed elation, pleasure, excitement, or energized positive affect (can include grandiose themes). \\
Hope-Optimism & Indicates expressed hopefulness, positive expectation, or belief in improvement or recovery. \\
Satisfaction-Contentment & Indicates expressed calm, peace, fulfillment, or a stable sense of well-being. \\
Neutral-None & Indicates no clearly expressed emotion, or a flat, matter-of-fact tone without affective coloring. \\
\midrule
\multicolumn{2}{c}{\textbf{Clinical Target: Behavioral Response}} \\
\midrule
Avoidance/Withdrawal & Behaviors aimed at escaping, avoiding, or disengaging from situations, people, or cues linked to distress or delusional beliefs. The individual reduces contact rather than taking action to increase safety. \\
Safety-Seeking/Protective Behaviors & Behaviors intended to prevent harm or increase perceived safety, while still engaging with the environment. These include active protective steps, precautionary rituals, or hypervigilant routines. \\
Confrontation/Resistance & Behaviors involving direct action toward the perceived threat or source of delusional content such as challenging, disrupting, or fighting back. \\
Help-Seeking & Behaviors involving reaching out to others including friends, family, clinicians to obtain support, validation, or practical assistance. \\
Self-Soothing/Regulation & Behaviors aimed at calming oneself, reducing distress, or managing emotions internally. These are coping behaviors rather than actions directed toward external threats. \\
Engagement/Acceptance & Behaviors showing constructive engagement with reality, acceptance of experiences, or efforts to refocus on meaningful activity. Often reflects cognitive reframing or functional coping. \\
Risky or Harmful Behaviors & Behaviors that pose risk of harm to the individual or others, including self-harm, aggression, reckless acts, or substance misuse. \\
Neutral/None & Indicates that no specific behavioral response is described, or the individual does not mention any actions related to their experiences. \\
\bottomrule
\end{tabular}
\caption{Comprehensive list of classes and their exact clinical definitions for Delusion Type, Affective Response, and Behavioral Response, as provided to the annotators and language models in the final guidelines.}
\label{tab:supp_class_definitions}
\end{table*}

\begin{table*}[htbp]
\centering
\small
\renewcommand{\arraystretch}{1.2}
\begin{tabular}{p{0.95\textwidth}}
\toprule
\textbf{Base Task Instructions and Output Formatting} \\ 
\midrule
\textbf{ANNOTATION GUIDELINES FOR CLINICAL TARGETS} \\
Audio diary transcripts will be annotated for various clinical targets, including hallucination content, delusions, substance use, and other relevant factors. \\
Each diary will be labeled as containing one or more clinical targets, and the relevant spans within the diary entry will be identified and annotated. These spans will represent triggers and entities associated with each clinical target, providing a detailed understanding of the individual's experiences and symptoms. \\
The following sections describe how each clinical target will be annotated. For each annotated clinical target, some entities are required and must always be present, while others may not always be present. \\
\\
\textbf{Note:} A transcript may contain multiple delusion themes. These can be: \\
$\bullet$ The same span with multiple themes (e.g., a statement that is both Persecutory and Reference) \\
$\bullet$ Different spans with different themes \\
List each \texttt{delusion\_span} and \texttt{delusion\_type} pair separately in order. \\
\\
Follow the following response template for your final answer. \\
\midrule
\textbf{Response Template and In-Context Formatting Examples} \\
\midrule
\textbf{\#\# example input} \\
I know they are monitoring my email, and I feel afraid all the time. \\
\textbf{\#\# response template} \\
\texttt{delusion\_span}: "I know they are monitoring my email" \\
\texttt{delusion\_type}: Persecutory \\
\texttt{affective\_span}: "I feel afraid all the time" \\
\texttt{affective\_category}: Fear-Anxiety \\
\texttt{affective\_intensity}: Moderate \\
\texttt{behavioral\_span}: null \\
\texttt{behavioral\_category}: null \\
\\
\textbf{\#\# example input (multiple delusion themes)} \\
God told me I am chosen to save humanity, and the government is trying to stop me. \\
\textbf{\#\# response template} \\
\texttt{delusion\_span}: "God told me I am chosen to save humanity" \\
\texttt{delusion\_type}: Religious \\
\texttt{delusion\_span}: "God told me I am chosen to save humanity" \\
\texttt{delusion\_type}: Grandiose \\
\texttt{delusion\_span}: "the government is trying to stop me" \\
\texttt{delusion\_type}: Persecutory \\
\texttt{affective\_span}: null \\
\texttt{affective\_category}: null \\
\texttt{affective\_intensity}: null \\
\texttt{behavioral\_span}: null \\
\texttt{behavioral\_category}: null \\
\\
\textbf{\#\# example input (no delusion detected)} \\
Today was a good day. I went to work and had lunch with a friend. \\
\textbf{\#\# response template} \\
\texttt{delusion\_span}: null \\
\texttt{delusion\_type}: null \\
\texttt{affective\_span}: null \\
\texttt{affective\_category}: null \\
\texttt{affective\_intensity}: null \\
\texttt{behavioral\_span}: null \\
\texttt{behavioral\_category}: null \\
\bottomrule
\end{tabular}
\caption{The high-level prompt structure (Base Task Instructions) provided to the language models. This foundational section establishes the extraction task, provides rules for handling multiple overlapping labels, and enforces a strict structured output format using explicit in-context examples.}
\label{tab:supp_annotation_instructions}
\end{table*}

\begin{table*}[htbp]
\centering
\small
\renewcommand{\arraystretch}{1.2}
\begin{tabular}{p{0.95\textwidth}}
\toprule
\textbf{Framework 1: Direct Adjudication Prompt (Example: Delusion Type)} \\ 
\midrule
You are an expert clinical judge evaluating delusion type annotations. Two models have annotated the same text and disagree on the delusion type. \\
\\
Here are the original annotation guidelines that were used: \\
\texttt{\{delusion\_guidelines\}} \\
\\
--- Original Text --- \\
\texttt{\{text\}} \\
\\
--- MODEL A ANALYSIS --- \\
Thinking Process: \texttt{\{glm\_thinking\}} \\
Detailed Annotation: \texttt{\{glm\_response\}} \\
Classification: \texttt{\{glm\_delusion\}} \\
\\
--- MODEL B ANALYSIS --- \\
Thinking Process: \texttt{\{gptoss\_thinking\}} \\
Detailed Annotation: \texttt{\{gptoss\_response\}} \\
Classification: \texttt{\{gptoss\_delusion\}} \\
\\
Based on the clinical guidelines above and considering both models' reasoning and annotation spans, determine the correct classification. You may: \\
- Choose Model A if their annotation is more accurate \\
- Choose Model B if their annotation is more accurate \\
- Choose Combined if both have valid points and the final answer should merge or differ from both \\
\\
Provide your judgment in this exact format: \\
WINNER: (Model A | Model B | Combined) \\
REASONING: (Brief explanation) \\
CORRECT\_TYPE: (The correct delusion type or comma-separated list if multiple) \\
\bottomrule
\end{tabular}
\caption{System prompt used for the Direct Adjudication framework. In this single-pass approach, the judge model is provided with the independent reasoning traces of both original annotators and asked to select or synthesize a final clinical label.}
\label{tab:supp_direct_prompt}
\end{table*}

\begin{table*}[htbp]
\centering
\small
\renewcommand{\arraystretch}{1.2}
\begin{tabular}{p{0.95\textwidth}}
\toprule
\textbf{Framework 2: Conversational Debate (Role Instructions \& Discussion Prompt)} \\ 
\midrule
\textbf{Annotator 1 Role:} You are Annotator 1. Review the disagreement with Annotator 2 and either: Defend your original annotation with clinical reasoning; Concede to Annotator 2's annotation if you find their reasoning more accurate; Propose a combined solution if both annotations have merit. Provide brief clinical reasoning focused on the annotation guidelines. \\
\\
\textbf{Annotator 2 Role:} You are Annotator 2. Review Annotator 1's response and either: Defend your original annotation with clinical reasoning... (same as Annotator 1). \\
\\
\textbf{Discussion Prompt:} \\
You are a professional clinical annotator participating in an annotation discussion to resolve disagreements on delusion type classification. \\
\texttt{\{delusion\_guidelines\}} \\
\\
Original text: \texttt{\{text\}} \\
\\
Annotation disagreement on: Delusion Type \\
Annotator 1's original annotation and reasoning: \\
- Delusion span: \texttt{\{glm\_delusion\_span\}} \\
- Delusion type: \texttt{\{glm\_delusion\_type\}} \\
- Original thinking: \texttt{\{glm\_thinking\}} \\
- Original response: \texttt{\{glm\_response\}} \\
\\
Annotator 2's original annotation and reasoning: \\
- Delusion span: \texttt{\{gpt\_delusion\_span\}} \\
- Delusion type: \texttt{\{gpt\_delusion\_type\}} \\
- Original thinking: \texttt{\{gpt\_thinking\}} \\
- Original response: \texttt{\{gpt\_response\}} \\
\\
\texttt{\{conversation\_history\}} \\
\texttt{\{current\_role\_instruction\}} \\
\midrule
\textbf{Framework 2: Conversational Debate (Final Adjudicating Judge Prompt)} \\
\midrule
You are an expert clinical judge resolving an annotation disagreement on delusion type classification. \\
\texttt{\{delusion\_guidelines\}} \\
\\
Original text: \texttt{\{text\}} \\
\\
\textit{(... Original Annotator 1 and Annotator 2 states injected here ...)} \\
\\
Discussion between annotators: \\
\texttt{\{conversation\_history\}} \\
\\
Based on the clinical guidelines, determine which annotator's ORIGINAL delusion type value is correct: \\
- If Annotator 1's original value (\texttt{\{glm\_delusion\_type\}}) is correct, choose Annotator 1 \\
- If Annotator 2's original value (\texttt{\{gpt\_delusion\_type\}}) is correct, choose Annotator 2 \\
- If neither is fully correct and a different/combined value is needed, choose Combined \\
\\
Respond in this format: \\
Winner: (Annotator 1 or Annotator 2 or Combined) \\
Final delusion\_type: (the correct value - must match winner's original value if Annotator 1 or 2, or new value if Combined) \\
Reasoning: (brief clinical justification) \\
\bottomrule
\end{tabular}
\caption{System prompts used for the Conversational Debate framework. This pipeline utilizes an iterative, multi-turn discussion between the original annotator agents, guided by role instructions, before the entire debate history is passed to a final judge model for adjudication.}
\label{tab:supp_debate_prompts}
\end{table*}

\begin{table*}[htbp]
\centering
\small
\begin{tabular}{lcccc}
\toprule
\textbf{Delusion Type} & \textbf{GPT-OSS} & \textbf{GLM-4.6} & \textbf{Qwen-3} & \textbf{Manual} \\ 
\midrule
Persecutory & 77 & 66 & 57 & 58 \\ 
None & 43 & 52 & 61 & 59 \\ 
Reference & 7 & 4 & 4 & 4 \\ 
Control & 4 & 3 & 2 & 2 \\ 
Unspecified & 1 & 3 & 1 & 3 \\ 
Nihilistic & 2 & 2 & 1 & 1 \\ 
Thought Broadcasting & 1 & 1 & 1 & 3 \\ 
Religious & 1 & 2 & 2 & 0 \\ 
Somatic & 1 & 2 & 1 & 1 \\ 
Mind Reading & 0 & 1 & 0 & 1 \\ 
Grandiosity & 0 & 0 & 2 & 0 \\ 
Thought Insertion & 1 & 0 & 0 & 0 \\ 
\bottomrule
\end{tabular}
\caption{Distribution of specific \textit{Delusion Type} labels in the test set extracted by the three primary models under the fully contextualized Level 4 prompt setting, compared against manual expert annotations.}
\label{tab:supp_delusion_distribution}
\end{table*}

\begin{table*}[htbp]
\centering
\small
\begin{tabular}{lcccc}
\toprule
\textbf{Affective Response} & \textbf{GPT-OSS} & \textbf{GLM-4.6} & \textbf{Qwen-3} & \textbf{Manual} \\ 
\midrule
Fear-Anxiety & 73 & 61 & 71 & 51 \\ 
Sadness-Despair & 25 & 26 & 25 & 11 \\ 
Anger-Frustration & 14 & 13 & 9 & 8 \\ 
Satisfaction-Contentment & 1 & 1 & 1 & 0 \\ 
Hope-Optimism & 0 & 1 & 0 & 0 \\ 
\bottomrule
\end{tabular}
\caption{Distribution of specific \textit{Affective Response} labels in the test set extracted by the three primary models under the fully contextualized Level 4 prompt setting, compared against manual expert annotations.}
\label{tab:supp_affective_distribution}
\end{table*}

\begin{table*}[htbp]
\centering
\small
\begin{tabular}{lcccc}
\toprule
\textbf{Behavioral Response} & \textbf{GPT-OSS} & \textbf{GLM-4.6} & \textbf{Qwen-3} & \textbf{Manual} \\ 
\midrule
Avoidance/Withdrawal & 36 & 32 & 31 & 24 \\ 
Help-Seeking & 18 & 13 & 20 & 1 \\ 
Safety-Seeking/Protective & 9 & 8 & 11 & 4 \\ 
Self-Soothing/Regulation & 6 & 3 & 6 & 1 \\ 
Confrontation/Resistance & 3 & 8 & 4 & 1 \\ 
Engagement/Acceptance & 2 & 8 & 2 & 1 \\ 
Risky/Harmful & 5 & 3 & 1 & 1 \\ 
\bottomrule
\end{tabular}
\caption{Distribution of specific \textit{Behavioral Response} labels in the test set extracted by the three primary models under the fully contextualized Level 4 prompt setting, compared against manual expert annotations.}
\label{tab:supp_behavioral_distribution}
\end{table*}

\begin{table*}[htbp]
\centering
\small
\begin{tabular}{ll ccc c}
\toprule
\textbf{Model} & \textbf{Prompt Level} & \textbf{Micro-averaged Precision} & \textbf{Micro-averaged Recall} & \textbf{Micro-averaged F1} & \textbf{Example F1} \\ 
\midrule
\multirow{4}{*}{\textbf{GLM-4.6}} 
 & Level 1 (Category)    & 0.586 & \textbf{0.890} & 0.707 & 0.773 \\ 
 & Level 2 (Definitions) & 0.575 & \textbf{0.890} & 0.699 & 0.743 \\ 
 & Level 3 (Rules)       & 0.567 & 0.808 & 0.667 & 0.711 \\ 
 & Level 4 (Full Prompt) & \textbf{0.726} & 0.836 & \textbf{0.777} & \textbf{0.811} \\ 
\midrule
\multirow{4}{*}{\textbf{GPT-OSS}} 
 & Level 1 (Category)    & 0.525 & 0.849 & 0.649 & 0.737 \\ 
 & Level 2 (Definitions) & 0.575 & \textbf{0.890} & 0.699 & 0.775 \\ 
 & Level 3 (Rules)       & 0.612 & 0.863 & 0.716 & 0.785 \\ 
 & Level 4 (Full Prompt) & \textbf{0.674} & 0.877 & \textbf{0.762} & \textbf{0.798} \\ 
\midrule
\multirow{4}{*}{\textbf{Qwen-3-235B}} 
 & Level 1 (Category)    & 0.598 & 0.753 & 0.667 & 0.722 \\ 
 & Level 2 (Definitions) & 0.598 & \textbf{0.795} & 0.682 & 0.771 \\ 
 & Level 3 (Rules)       & 0.734 & \textbf{0.795} & \textbf{0.763} & \textbf{0.811} \\ 
 & Level 4 (Full Prompt) & \textbf{0.746} & 0.726 & 0.736 & 0.793 \\ 
\bottomrule
\end{tabular}
\caption{Comprehensive performance breakdown for \textit{Delusion Type} extraction against human expert annotations.}
\label{tab:app_delusion_metrics}
\end{table*}

\begin{table*}[htbp]
\centering
% \resizebox{1.0\linewidth}{!}{%
\small
\begin{tabular}{llcccc}
\toprule
\multirow{2}{*}{\textbf{Clinical Target}} & \multirow{2}{*}{\textbf{Prompt Level}} & \multicolumn{4}{c}{\textbf{Micro-averaged F1}} \\
\cmidrule(lr){3-6}
& & \textbf{GLM-4.6} & \textbf{GPT-OSS} & \textbf{Qwen-3-235B} & \textbf{Majority Vote} \\
\midrule
\multirow{4}{*}{\textbf{Delusion Presence}} 
& Level 1 (Category)    & 0.865 & 0.855 & 0.809 & 0.853 \\
& Level 2 (Definitions) & 0.836 & 0.871 & 0.837 & 0.868 \\
& Level 3 (Rules)       & 0.822 & 0.886 & 0.868 & 0.878 \\
& Level 4 (Full Prompt) & 0.857 & 0.873 & 0.839 & 0.872 \\
\midrule
\multirow{4}{*}{\textbf{Delusion Type}} 
& Level 1 (Category)    & 0.707 & 0.649 & 0.667 & 0.746 \\
& Level 2 (Definitions) & 0.699 & 0.699 & 0.682 & 0.715 \\
& Level 3 (Rules)       & 0.667 & 0.716 & 0.763 & 0.790 \\
& Level 4 (Full Prompt) & 0.777 & 0.762 & 0.736 & 0.779 \\
\midrule
\multirow{4}{*}{\textbf{Affective Response}} 
& Level 1 (Category)    & 0.797 & 0.774 & 0.813 & 0.839 \\
& Level 2 (Definitions) & 0.767 & 0.803 & 0.732 & 0.780 \\
& Level 3 (Rules)       & 0.730 & 0.803 & 0.750 & 0.797 \\
& Level 4 (Full Prompt) & 0.694 & 0.760 & 0.768 & 0.800 \\
\midrule
\multirow{4}{*}{\textbf{Behavioral Response}} 
& Level 1 (Category)    & 0.507 & 0.456 & 0.513 & 0.545 \\
& Level 2 (Definitions) & 0.519 & 0.545 & 0.500 & 0.564 \\
& Level 3 (Rules)       & 0.458 & 0.528 & 0.500 & 0.482 \\
& Level 4 (Full Prompt) & 0.615 & 0.582 & 0.582 & 0.640 \\
\bottomrule
\end{tabular}%
% }
\caption{Comprehensive performance comparison (Micro-averaged F1) across varying prompt complexities for individual LLMs and the Tri-Model Majority Vote.}
\label{tab:supp_prompt_performance}
\end{table*}

\begin{table*}[htbp]
\centering
\small
% \resizebox{1.0\linewidth}{!}{%
\begin{tabular}{l cc ccccc}
\toprule
& \multicolumn{2}{c}{\textbf{Agreement Subset ($N=91$)}} & \multicolumn{5}{c}{\textbf{Disagreement Subset ($N=31$)}} \\
\cmidrule(lr){2-3} \cmidrule(lr){4-8}
\textbf{Clinical Target} & \textbf{GLM/GPT Consensus} & \textbf{Qwen-3} & \textbf{GLM-4.6} & \textbf{GPT-OSS} & \textbf{Qwen-3} & \textbf{Direct Judge} & \textbf{Debate} \\ 
\midrule
Delusion Presence   & 0.894 & 0.847 & 0.769 & 0.833 & 0.821 & 0.800 & 0.667 \\ 
Delusion Type       & 0.857 & 0.800 & 0.615 & 0.603 & 0.612 & 0.618 & 0.545 \\ 
Affective Response  & 0.850 & 0.800 & 0.409 & 0.612 & 0.711 & 0.590 & 0.500 \\ 
Behavioral Response & 0.667 & 0.375 & 0.606 & 0.567 & 0.635 & 0.571 & 0.629 \\ 
\bottomrule
\end{tabular}%
% }
\caption{Micro-averaged F1 performance stratified by initial annotator agreement. The Agreement Subset ($N=91$) compares the natural consensus of GLM-4.6 and GPT-OSS against Qwen-3 as an independent baseline. All evaluations used the Level 4 prompt.}
\label{tab:supp_adjudication_stratified}
\end{table*}

\begin{table*}[htbp]
\centering
\small
\begin{tabular}{lcccccc}
\toprule
\multirow{2}{*}{\textbf{Clinical Target}} & \multicolumn{2}{c}{\textbf{GLM vs GPT}} & \multicolumn{2}{c}{\textbf{GPT vs Qwen}} & \multicolumn{2}{c}{\textbf{GLM vs Qwen}} \\
\cmidrule(lr){2-3} \cmidrule(lr){4-5} \cmidrule(lr){6-7}
& \textbf{Micro} & \textbf{Macro} & \textbf{Micro} & \textbf{Macro} & \textbf{Micro} & \textbf{Macro} \\
\midrule
Delusion Presence & 0.709 & 0.709 & 0.672 & 0.672 & 0.721 & 0.721 \\
Delusion Type & 0.777 & 0.515 & 0.755 & 0.481 & 0.731 & 0.433 \\
Affective Response & 0.646 & 0.295 & 0.694 & 0.412 & 0.648 & 0.401 \\
Behavioral Response & 0.239 & 0.092 & 0.424 & 0.182 & 0.300 & 0.196 \\
\bottomrule
\end{tabular}
\caption{Pairwise agreement (Cohen's $\kappa$) across the three primary LLMs using the fully contextualized Level 4 prompt. Micro-averaged $\kappa$ is computed on flattened binary indicator matrices; macro-averaged $\kappa$ is the unweighted mean of per-class $\kappa$ values. For \textit{Delusion Presence} (binary), the two are equivalent.}
\label{tab:model_agreement_kappa}
\end{table*}

\begin{table*}[htbp]
\centering
\small
\begin{tabular}{l cc cc cc}
\toprule
& \multicolumn{2}{c}{\textbf{GLM-4.6 vs GPT-OSS}} & \multicolumn{2}{c}{\textbf{GPT-OSS vs Qwen-3}} & \multicolumn{2}{c}{\textbf{GLM-4.6 vs Qwen-3}} \\
\cmidrule(lr){2-3} \cmidrule(lr){4-5} \cmidrule(lr){6-7}
\textbf{Clinical Target} & \textbf{Agree} & \textbf{Disagree} & \textbf{Agree} & \textbf{Disagree} & \textbf{Agree} & \textbf{Disagree} \\ 
\midrule
Delusion Presence & 86.1\% & 13.9\% & 83.6\% & 16.4\% & 86.1\% & 13.9\% \\ 
Delusion Type & 75.4\% & 24.6\% & 73.8\% & 26.2\% & 76.2\% & 23.8\% \\ 
Affective Response & 65.6\% & 34.4\% & 66.4\% & 33.6\% & 65.6\% & 34.4\% \\ 
Behavioral Response & 47.5\% & 52.5\% & 58.2\% & 41.8\% & 53.3\% & 46.7\% \\ 
\bottomrule
\end{tabular}
\caption{Baseline pairwise agreement rates across the three primary LLMs using the fully contextualized Level 4 prompt. Agreement represents exact set matches; disagreement includes both partial overlap and zero intersection.}
\label{tab:model_agreement_original}
\end{table*}
\end{document}